# Rethinking Tokenization for Rich Morphology: The Dominance of Unigram over BPE and Morphological Alignment

**Saketh Reddy Vemula**[1] **Sandipan Dandapat**[2]
**Dipti Misra Sharma**[1] **Parameswari Krishnamurthy**[1]
[1]IIIT Hyderabad, India [2]Microsoft R&D, India
saketh.vemula@research.iiit.ac.in sadandap@microsoft.com
{dipti, param.krishna}@iiit.ac.in

## Abstract

The relationship between tokenizer algorithm (e.g., Byte-Pair Encoding (BPE), Unigram), morphological alignment, tokenization quality (e.g., compression efficiency), and downstream performance remains largely unclear, particularly for languages with complex morphology. In this paper, we conduct a comprehensive evaluation of tokenizers using small-sized BERT models—from pre-training through fine-tuning—for Telugu (agglutinative), along with preliminary evaluation in Hindi (primarily fusional with some agglutination) and English (fusional). To evaluate morphological alignment of tokenizers in Telugu, we create a dataset containing gold morpheme segmentations of 600 derivational and 7000 inflectional word forms.

Our experiments reveal two key findings for Telugu. First, the choice of tokenizer algorithm is the most significant factor influencing performance, with Unigram-based tokenizers consistently outperforming BPE across most settings. Second, while better morphological alignment shows a moderate, positive correlation with performance on text classification and structure prediction tasks, its impact is secondary to the tokenizer algorithm. Notably, hybrid approaches that use morphological information for pre-segmentation significantly boost the performance of BPE, though not Unigram. Our results further showcase the need for comprehensive intrinsic evaluation metrics for tokenizers that could explain downstream performance trends consistently.

## 1 Introduction

Modern natural language processing (NLP) tools suffer from systematic performance bias towards high-resource languages, thereby affecting the performance in low-resource languages (Joshi et al., 2020; Aji et al., 2022; Levy et al., 2023; Ramesh et al., 2023). Although large language models (LLMs) have revolutionized NLP by delivering state-of-the-art performances across a wide range of tasks (Qin et al., 2024), they, however, owe their success not only to scaling but also foundational decisions—such as tokenizer choice (Ahuja et al., 2022; Rust et al., 2020). Recent efforts toward building a more inclusive and equitable NLP ecosystem include the creation of large-scale resources (Kakwani et al., 2020a; Ramesh et al., 2022), as well as developing key architectural innovations, methodological insights, and frameworks for fairer evaluation in low-resource and morphologically complex languages (Choudhury, 2023).

Language-specific processing for languages with considerably different morphological typologies has become increasingly relevant while developing small-scale models (Khanuja et al., 2021; Dabre et al., 2022). As we shift towards building efficient and compact language models, particularly for low-resource settings, incorporating linguistic cues—such as morphology and syntactic features—would become crucial for improving their performance and generalizability (Wiemerslage et al., 2022).

Morphologically complex and agglutinative languages present us with one such opportunity. These languages typically exhibit a large number of surface forms per lemma due to the agglutination or fusion of multiple grammatical markers—such as tense, number, case, and person—onto a single root (Comrie, 1989; Haspelmath and Sims, 2013). This morphological richness results in a high type-to-token ratio, contributing to data sparsity and making such languages harder to model effectively (Cotterell et al., 2018). For instance, agglutinative languages tend to have longer words and more unique word forms due to words being composed of many individual morphemes (Ramasamy et al., 2012). Subword tokenizers that generate semantically meaningless segments (Beinborn and Pinter, 2023; Libovický and Helcl, 2024) fails to handle this complexity, thereby producing suboptimal performance (Batsuren et al., 2024). Whether a morphologically informed approach to tokenization would better handle such grammatical complexity and improve the downstream performance remains debated.

In this work, we focus on the following question: *does morphologically aligned approaches to tokenization better handle the complexity of morphologically complex languages?* To comprehensively evaluate this, we explore a range of tokenization approaches with varying levels of granularity and incorporate different techniques for aligning token boundaries with morphological structure. For each tokenizer variant, we pre-train, fine-tune and evaluate encoder-only models with BERT (Devlin et al., 2019) architecture at 8.5 million parameter (excluding parameters count in embedding layer) scale on various benchmarks. We focus on Telugu due to its highly agglutinative and complex word formation. We perform similar evaluations in Hindi and English for evaluating whether similar trends are observed in comparatively less complex languages. Upon observing consistent differences in downstream performance, we test and discuss two competing hypotheses that could explain those trends:

1. **Morphological Alignment**: Morphologically aligned tokenizer capture more semantically meaningful tokens which lead to improved modeling and performance.
2. **Tokenization Quality**: Tokenizer with higher compression efficiency or better distribution of token frequencies lead to improved modeling and performance.

To test morphological alignment in Telugu, we adapt existing morphological analyzers and create a dataset containing gold morpheme segmentations for both inflectional and derivational word forms.

## 2 Related Work

Tokenization has been a fundamental preprocessing step in all modern NLP systems, including large language models (LLMs). Popular approaches include subword tokenization algorithms such as Byte-Pair Encoding (BPE) (Gage, 1994; Shibata et al., 2000; Sennrich et al., 2016), the Unigram Language Model (ULM) (Kudo, 2018), and WordPiece (Schuster and Nakajima, 2012). Several improvements have been proposed following these methods, aiming either to produce more statistically effective tokens (Kudo and Richardson, 2018) or to align tokens with morpheme boundaries (Libovický and Helcl, 2024; Zhu et al., 2024; Creutz and Lagus, 2007; Smit et al., 2014).

Evaluating tokenizers intrinsically include various approaches. Some of them are measuring compression efficiency (Schmidt et al., 2024; Zouhar et al., 2023), cognitive plausibility (Beinborn and Pinter, 2023), and morphological alignment (Batsuren et al., 2024; Uzan et al., 2024). However, no single evaluation method has emerged that reliably explains tokenizer quality or correlates well with extrinsic performance on downstream tasks (Cognetta et al., 2024; Chizhov et al., 2024; Goldman et al., 2024; Ali et al., 2024; Reddy et al., 2025).

Morphologically aligned tokenization has been argued to enhance language understanding and improve downstream performance of language models (Hou et al., 2023; Fujii et al., 2023; Jabbar, 2024; Batsuren et al., 2024; Truong et al., 2024; Asgari et al., 2025). However, many of these works prematurely equate improvements in language modeling with lower training loss—as measured by perplexity—or faster convergence. Additionally, most studies have been limited to high-resource and morphologically less complex languages such as English.

## 3 Evaluating Tokenization Approaches

To comprehensively evaluate the effect of different tokenization approaches, we adopt a multi-stage experimental framework. We evaluate each language model trained using a tokenizer variant on a diverse set of downstream tasks and keep all the hyperparameters strictly constant across the variants of tokenizers in order to isolate the tokenizer's effect on language modeling. We train both tokenizers and languages models on WMT News Crawl corpus[1] (Chelba et al., 2014). We randomly choose a subset of 10 million sentences for each language from the corpus. For Telugu, the corpus did not provide the desired volume of data. Therefore, we add additional sentences from IndicCorp (Kunchukuttan et al., 2020) dataset to meet the target size. We ensure no duplication of sentences during this process. Refer Appendix A.1 for corpus statistics.

Figure 1 presents our methodology to evaluate various tokenizer variants. For each language, we systematically vary the tokenization strategy by employing tokenizers at different linguistics levels. Namely, we include character-, subword-, *hybrid*-, *morphemic*-, and word-level tokenizers. Character-level and Word-level tokenizers provide us with two extremes in granularity of tokens generated. Subword-level tokenizers (i.e., BPE and Unigram) were trained on the raw WMT corpora. To approximate a linguistic morpheme and at the same time limit vocabulary size, we combine unsupervised morphological segmenter, Morfessor (Creutz and Lagus, 2007; Smit et al., 2014), or morphological analyzer (Rao et al., 2011) with subword approaches to create *hybrid* tokenizers. Initial segmentation was performed with Morfessor or morphological analyzers to create an intermediate morphologically pre-segmented corpus (as shown in Figure 1) and later subword tokenizer was trained on top of the segmented text. We refer to the word-level tokenizer trained on the morphologically pre-segmented corpus as *morphemic-level* tokenizer. Out-of-vocabulary (OOV) words were handled using a special unknown token (`[UNK]`) in case of *morphemic-* and word-level tokenizers. Note that, we strictly restrict the vocabulary sizes to predefined limit across all the variants in order to provide a fair comparison. We also vary these vocabulary sizes across subword and hybrid variants.

[1] https://data.statmt.org/news-crawl/

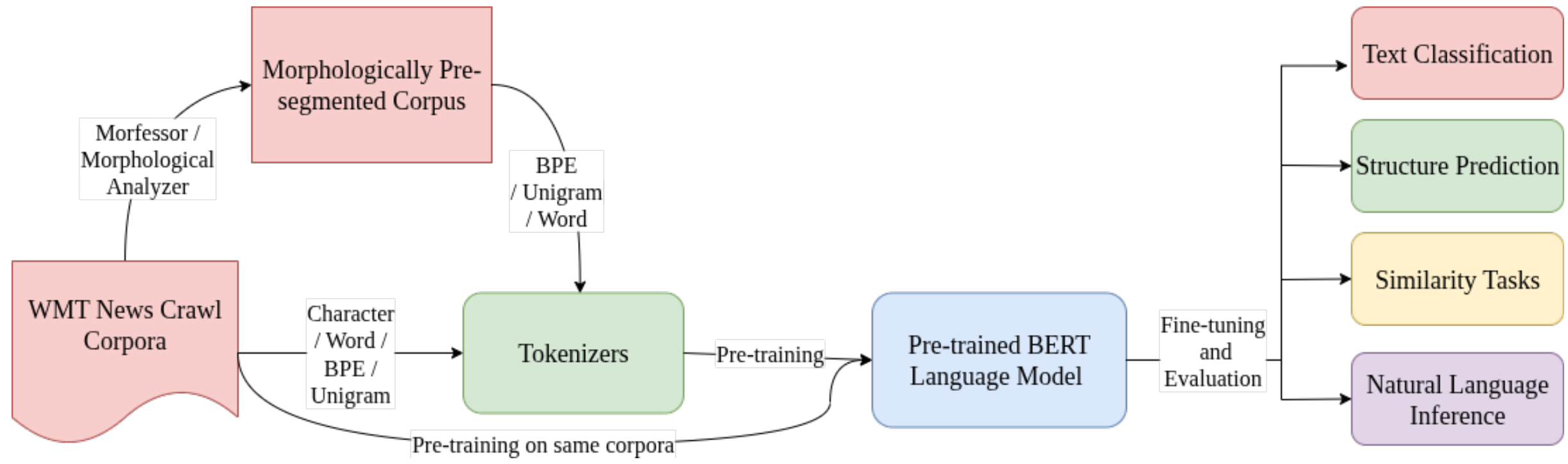

Figure 1: Our methodology for evaluating the effect of morphological alignment of tokenizers on language modeling.

### 3.1 Experimental Settings

We choose encoder-only transformer (Vaswani et al., 2023) model with standard BERT (Bidirectional Encoder Representations from Transformers) (Devlin et al., 2019) architecture across all our experiments. We evaluate and observe language understanding capabilities of these models while varying the tokenizer. In total, we pre-train 72 models at 8.5 million parameter scale (excluding parameters in embedding layer) across all the variants. All the models were pre-trained on an NVIDIA RTX 6000 GPU with 50 GB VRAM and later fine-tuned on 4 NVIDIA GeForce RTX 2080 GPUs. Each model was trained for 175,000 steps with 16-bit precision. Hyperparameters choice for pre-training are listed in table 11 in Appendix.

| Task Name | Telugu | Hindi | English |
|---|---|---|---|
| ***Text Classification*** | | | |
| Sentiment Analysis | ✓ | ✓ | ✓ |
| Discourse Mode | ✗ | ✓ | ✗ |
| Intent Classification | ✓ | ✗ | ✗ |
| Word & Definition | ✗ | ✗ | ✓ |
| Word & Morphology | ✗ | ✗ | ✓ |
| ***Structure Prediction*** | | | |
| POS Tagging | ✓ | ✓ | ✓ |
| NER | ✓ | ✓ | ✓ |
| Dependency Parsing | ✓ | ✓ | ✓ |
| ***Similarity Assessment*** | | | |
| Paraphrase Detection | ✓ | ✓ | ✗ |
| Sentence Similarity | ✗ | ✓ | ✓ |
| Word & Word | ✓* | ✓* | ✓ |
| ***Natural Language Inference*** | | | |
| NLI | ✓ | ✓ | ✓ |

Table 1: Downstream tasks considered for evaluation, grouped into categories. "✓" denotes the presence while "✗" denotes absence of a task for the language. "*" marks datasets curated in this work (cf. Appendix A.3).

To evaluate these models, we utilize available benchmarks that encompass diverse set of downstream tasks. Table 1 presents an overview of tasks included in our evaluation. We evalute the pre-trained models on tasks from benchmarks such as GLUE (Wang et al., 2019), IndicGLUE (Kakwani et al., 2020b), and IndicXTREME (Doddapaneni et al., 2023). These tasks span diverse set of categories such as classification, structure prediction, similarity assessment, and natural language inference. We also include additional tasks such as in Batsuren et al. (2024) for English, and curate similar datasets in Hindi and Telugu (refer Appendix A.3 for details), to specifically test out-of-vocabulary generalization of tokenizers. Details and description related to each task and hyperparameters used while fine-tuning for each can be found in Appendix A.3 and Appendix A.5 respectively. For each task-variant combination, we perform three independent runs and report the mean and standard deviation to ensure robustness in downstream evaluation. We make the scripts used for pre-training and fine-tuning public: rethinking-tokenization-for-rich-morphology.

| Tokenizer | Pre-Tokenizer | **Text Classification** | | | **Structure Prediction** | | | **Similarity Assessment** | | | ***Overall Trend*** | | |
|---|---|---|---|---|---|---|---|---|---|---|---|---|---|
| *Vocabulary Size* | | **8192** | **16384** | **50277** | **8192** | **16384** | **50277** | **8192** | **16384** | **50277** | **8192** | **16384** | **50277** |
| Character | None (Naive) | 66.69 | 67.23 | 69.62 | 71.52 | 70.63 | 71.86 | 61.30 | 60.68 | 59.58 | 66.02 | 64.49 | 65.65 |
| BPE | None (Naive) | 69.39 | 66.44 | 68.57 | 71.33 | 71.14 | 71.22 | 63.28 | 62.74 | **62.64** | 66.40 | 64.95 | 65.78 |
| | Morfessor | **72.25** | **70.31** | **70.25** | 72.52 | **72.26** | **72.74** | 63.39 | **62.96** | 62.57 | **68.16** | **67.11** | **67.29** |
| | Morph Analyzer | 70.58 | 69.42 | 68.31 | **72.97** | 72.04 | 70.40 | **63.68** | 62.75 | 62.15 | 67.70 | 66.68 | 65.58 |
| *Vocabulary Average* | | 70.74 | 68.72 | 69.04 | 72.28 | 71.81 | 71.45 | 63.45 | 62.82 | 62.45 | 68.45 | 66.25 | 66.22 |
| Unigram | None (Naive) | 77.71 | **80.06** | **81.56** | 79.23 | **81.07** | **83.01** | 62.22 | **64.11** | **67.25** | 73.32 | **74.83** | **77.29** |
| | Morfessor | **78.98** | 79.78 | 81.16 | **79.60** | 80.03 | 82.27 | 63.28 | 63.77 | 64.24 | 73.72 | 74.09 | 75.64 |
| | Morph Analyzer | 78.90 | 79.49 | 78.96 | 79.45 | 80.62 | 81.58 | **65.75** | 63.57 | 63.02 | **73.93** | 74.39 | 74.27 |
| *Vocabulary Average* | | 78.53 | 79.78 | 80.56 | 79.43 | 80.58 | 82.29 | 63.75 | 63.82 | 64.83 | 73.66 | 74.44 | 75.73 |
| Word | None (Naive) | 68.90 | **70.21** | **74.00** | **71.52** | 71.52 | 77.11 | 56.00 | 57.56 | 57.82 | **66.04** | **66.41** | **70.40** |
| | Morfessor | 68.68 | 69.55 | 73.48 | 70.59 | **71.72** | **78.59** | 56.06 | **58.30** | 57.68 | 65.26 | 66.21 | 70.19 |
| | Morph Analyzer | **68.94** | 67.00 | 69.12 | 69.89 | 69.85 | 75.91 | **57.66** | 56.36 | **60.08** | 63.87 | 63.89 | 67.79 |
| *Vocabulary Average* | | 67.63 | 68.92 | 72.20 | 70.67 | 71.03 | 77.20 | 56.58 | 57.41 | 58.53 | 65.06 | 65.50 | 69.46 |

Table 2: Downstream performance of language models trained using different variants of tokenizer in Telugu. *Vocabulary Average* is the average of the scores across a vocabulary size (e.g., 8192) while varying tokenizer variant. *Overall Trend* presents an average score across all tasks showcasing high-level trends. For a vocabulary size, the best performing pre-tokenizer for a variant of tokenizer is **bolded**, while the best variant across all the combinations of pre-tokenizer and tokenizer is underlined. Best performing vocabulary size for a category of task and tokenizer combination is underlined with wavy.

### 3.2 Results and Observations

Table 2 presents the summarized downstream performance results for Telugu. Tasks are organized into three categories following table 1. Text classification reports the average accuracies across sentiment analysis, intent classification, and similarity classification tasks. Structure prediction includes the average F1-scores for part-of-speech tagging, named entity recognition, and the labeled attachment score for dependency parsing. Similarity assessment report average accuracies on paraphrase detection and word-level similarity classification. Detailed results for all languages and individual task scores are presented in Appendix A.6.

We observe that the naive[2] Unigram tokenizers consistently delivers the best overall performance with significant margins across most tokenizer variants and downstream tasks. Interestingly, the performance gains from naive BPE tokenizer were marginal compared to character- or word-level tokenizers. For text classification, naive BPE tokenizer performed worse than other approaches at larger vocabulary sizes. However, incorporating linguistically motivated strategies—particularly pre-segmentation using Morfessor—led to substantial improvements within the BPE framework. Hybrid approaches combining Morfessor and BPE outperformed their naive counterparts, with significant gains in both text classification and structure prediction tasks. Similar gains were not observed across vocabulary sizes in case of hybrid tokenizers involving Unigram framework. Only at smaller vocabulary sizes, do these tokenizers outperform their naive counterparts.

Furthermore, we observe consistent patterns with the optimal vocabulary sizes for different tokenizers. BPE performed best at lower vocabulary sizes, whereas Unigram achieved peak performance at higher vocabulary sizes. Additionally, the improvements from linguistically informed approaches were more consistent at smaller vocabulary sizes for tokenizers involving BPE and Unigram framework.

Following the analysis in Arnett and Bergen (2024), we test two competing hypothesis that could explain our observations: Morphological Alignment and Tokenization Quality. Note that since we did not include tokenizer variants involving morphological analyzers as pre-tokenizer for Hindi and English, we had only 12 data points (as compared to 18 data points in Telugu), thereby compromising the statistical power of many complex tests for those languages. For instance, we could perform correlation tests using fixed effects models only for Telugu. Therefore, the findings involving fixed effects models in further sections are only in Telugu and must be treated as exploratory and preliminary, not conclusive.

[2]We refer to tokenizers trained directly on the corpus without morphological pre-segmentation or pre-tokenization as "naive".

| Word | Pre-tokenizer | Tok | Segmentation | Gold | Pred | Recall | Precision |
|---|---|---|---|---|---|---|---|
| ఆధారపడతాము | *gold reference* | | ఆధారపడ + తా + ము | | | | |
| | - | BPE | ఆధార + పడ + తాము | [6, 8] | [4, 6] | 0.5 | 0.5 |
| | Morfessor | BPE | ఆధారపడ + తాము | [6, 8] | [6] | 0.5 | 1.0 |
| | Morph Analyzer | BPE | ఆధార + పడ + తాము | [6, 8] | [4, 6] | 0.5 | 0.5 |
| | - | UNI | ఆధారపడ + తాము | [6, 8] | [6] | 0.5 | 1.0 |
| | Morfessor | UNI | ఆధారపడ + తాము | [6, 8] | [6] | 0.5 | 1.0 |
| | Morph Analyzer | UNI | ఆధారపడ + తాము | [6, 8] | [6] | 0.5 | 1.0 |
| ఆర్థికాభివృద్ధికి | *gold reference* | | ఆర్థికాభివృద్ధి + కి | | | | |
| | - | BPE | ఆర్థి + కా + భివృద్ధి + కి | [15] | [5, 7, 15] | 1.0 | 0.33 |
| | Morfessor | BPE | ఆర్థ + ి + కా + భివృద్ధి + కి | [15] | [4, 5, 7, 15] | 1.0 | 0.25 |
| | Morph Analyzer | BPE | ఆర్థి + కా + భివృద్ధి + కి | [15] | [5, 7, 15] | 1.0 | 0.33 |
| | - | UNI | ఆర్థిక + ాభివృద్ధి + కి | [15] | [6, 15] | 1.0 | 0.5 |
| | Morfessor | UNI | ఆర్థిక + ాభివృద్ధి + క + ి | [15] | [6, 15, 16] | 1.0 | 0.33 |
| | Morph Analyzer | UNI | ఆర్థిక + ాభివృద్ధి + కి | [15] | [6, 15] | 1.0 | 0.5 |

Table 3: Example word forms in Telugu along with their MorphScores and segmentations produced by different tokenizers. "Gold" indicates the character-level morpheme boundary positions from the ground-truth annotations, while "Pred" shows the corresponding predicted boundary positions generated by each tokenizer variant. BPE denotes Byte-Pair Encoding tokenizer and UNI denotes Unigram tokenizer.

## 4 Morphological Alignment

One possible explanation for our observations is that morphologically aligned tokenization produced more meaningful tokens, which ultimately lead to improved language modeling and downstream performance. This explanation becomes even more compelling in the case of morphologically rich languages. In such languages, words are often formed by combining multiple morphemes, each carrying a distinct grammatical feature. It is therefore intuitive to assume that a tokenizer which explicitly segments these morphemes can generate more meaningful embeddings, thereby enhancing language modeling performance.

To evaluate this hypothesis, we utilize the existing boundary-based evaluation metric—MorphScore (Arnett and Bergen, 2024; Arnett et al., 2025)—for evaluating morphological alignment. Refer Appendix A.2 for detailed description of MorphScore. For Telugu, we create a dataset containing gold morpheme segmentations for approximately 600 derivational and 7000 inflectional words. To the best of our knowledge, this is the first dataset containing gold morpheme segmentations in Telugu. For Hindi and English, we utilize the existing dataset created in Arnett et al. (2025).

### 4.1 Morpheme Segmentations in Telugu

To evaluate morphological alignment in Telugu, we required gold morpheme segmentations that represent the ground-truth for a morphemic-segmentation (i.e., segmentation of a complex word form where each segment is semantically meaningful). We utilize existing Telugu morphological analyzer (Rao et al., 2011) and extract word forms that contain derivational and inflectional suffixes from paradigms. In total, we could extract 1297 derivational and 9275 inflectional unique word forms. We filter out word forms for which the segments, as analyzed by the morphological analyzer, do not combine to form the original word form. This is crucial as tokenizers simply segments complex words and does not transform the existing stem into its lemma. 634 derivational and 7458 inflectional unique word forms remained after filtering. These word forms along with their segmented outputs serve as our gold morpheme segmentations. We further validate the correctness of the segmentations manually and found no considerable errors. We make the dataset public: TeluguMorphScore.

### 4.2 Evaluation

MorphScore assesses how well segmentations from tokenizer correspond to ground-truth morphological boundaries. The algorithm operates by comparing character-level boundary positions between gold morphological segmentations and tokenizer outputs, computing recall and precision metrics to quantify morphological alignment. Refer Appendix A.2 for an example walkthrough. Words tokenized as a single token are excluded from the evaluation in order to consider only complex word forms into the final score. Similarly, words with no ground-truth morpheme boundaries (i.e., word consisting of a single morpheme) are also excluded. The inclusion of both recall and precision metrics offers insights into whether a tokenizer tend toward over-segmentation or under-segmentation relative to morphological boundaries. Approximately 2000 word forms were evaluated consistently across all variants of tokenizers after all exclusions. Table 3 lists out few example word forms in Telugu along with the morphological alignment scores calculated for each variant of tokenizer at vocabulary size of 16384.

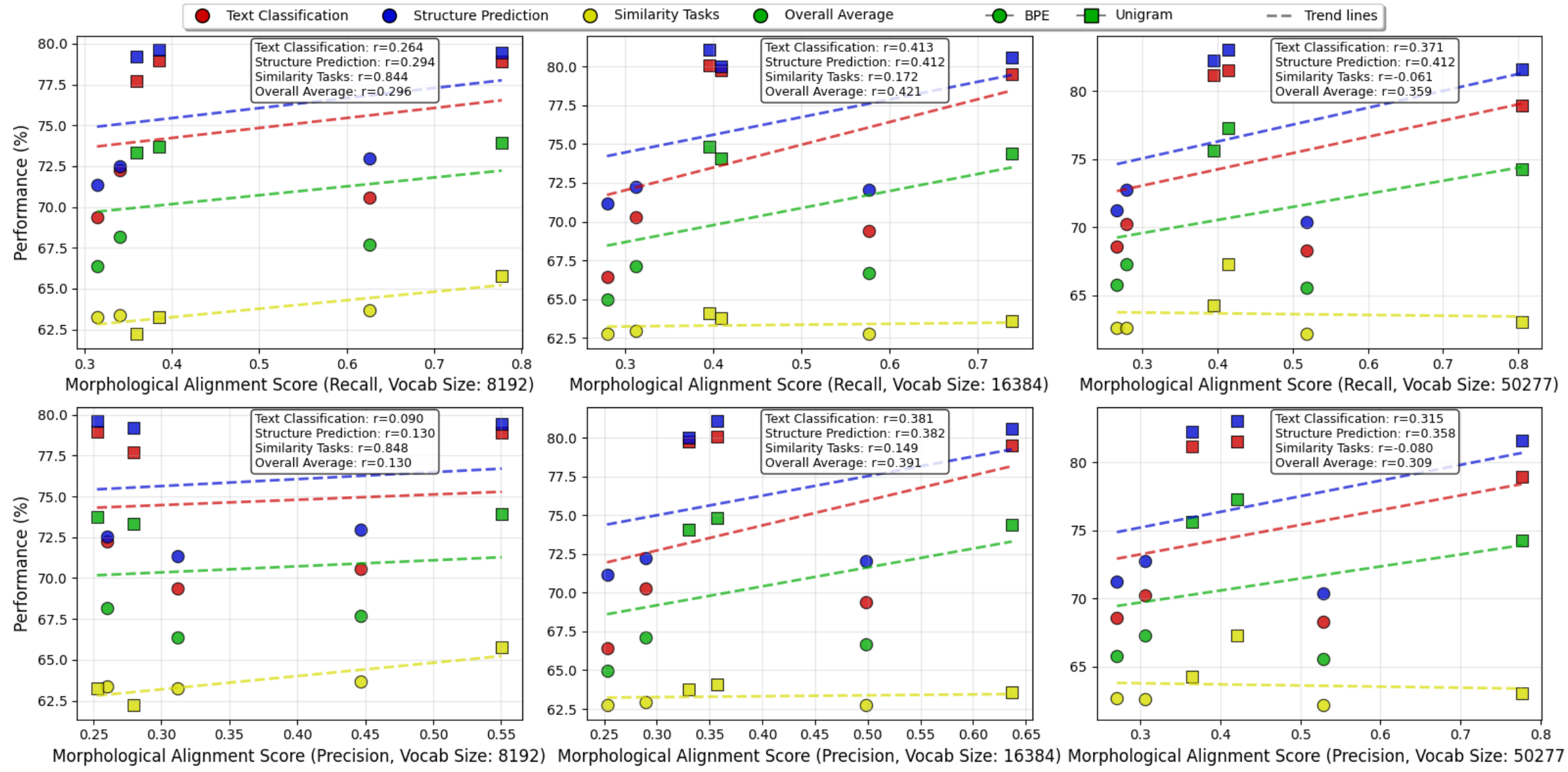

Figure 2: Variation in Downstream Performance with Morphological Alignment Scores for different tokenizer variants in Telugu, grouped by Vocabulary Size.

Figure 2 presents dot plots grouped by vocabulary sizes illustrating the relationship between morphological alignment scores and downstream performance for Telugu. Figure 5 in Appendix A.2 shows comprehensive plot combining all vocabulary sizes. The corresponding plots for Hindi (Figure 6) and English (Figure 7) are also included in Appendix A.2. Tables 5, 6 and 7 includes detailed morphscores for each language in Appendix A.2.

### 4.3 Results & Discussion

Based on our analysis, we found that there is a statistically significant but moderate positive correlation between the morphological alignment of a tokenizer and its performance on downstream tasks. However, we observe that the choice of tokenizer algorithm (BPE vs. Unigram) has a much stronger impact on performance than morphological alignment alone.

Initially, we explored the direct relationship between morphological metrics (such as recall, precision, and F1-score) and downstream task performance. Pearson correlation to account for linear relationship showed weak to moderate positive correlations. For example, the correlation between the overall trend and morphological F1-score was not statistically significant ($r = 0.332$, $p = 0.179$). This indicates the absence of a strong linear relationship. Spearman correlation, on the other hand, which accounts for monotonic relationship, revealed a stronger and more significant relationship. For instance, correlation between overall trend and recall was $0.486$ ($p = 0.041$), and with F1-score it was $0.474$ ($p = 0.047$). The strongest correlation among task categories was observed with structure prediction ($r = 0.478$, $p = 0.045$ for recall).

We performed ANOVA (Analysis of Variance) and ANCOVA (Analysis of Covariance) tests to disentangle the effects of different factors. Across almost all tasks, tokenizer (BPE vs. Unigram) had a very large and statistically significant effect on performance. For instance, in the two-way ANOVA for text classification, the F-statistic for C(Tokenizer) was $276.82$ ($p < 0.001$), indicating that it is a pri-mary driver of performance difference. On the other hand, the pre-tokenizer showed no significant effect on performance. When ANCOVA test introduced the F1-score (from MorphScores) as a covariate, for structure prediction, even after accounting for the powerful effects of the pre-tokenizer and tokenizer, it remained a statistically significant predictor of performance ($F = 5.71$, $p = 0.033$). Thus better morphological alignment independently contributes to better performance on syntax-based tasks. This effect was not significant for other tasks once tokenizer choice was factored in.

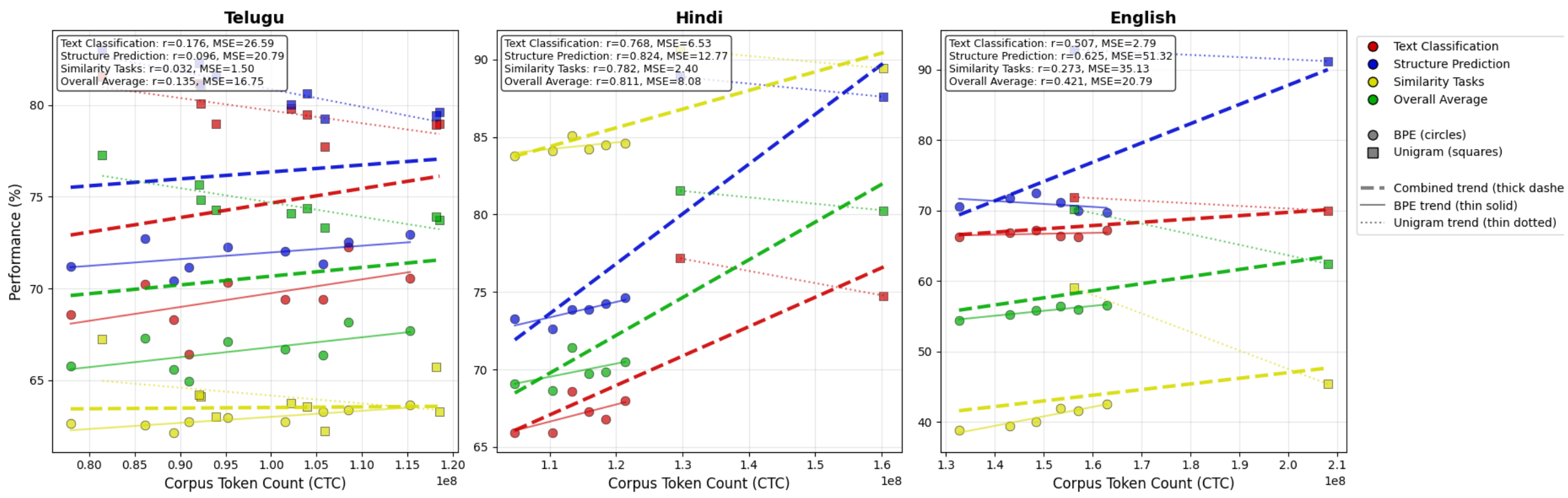

Figure 3: Variation in Downstream Performance with Corpus Token Count (CTC) for different tokenizer variants in Telugu, Hindi, and English.

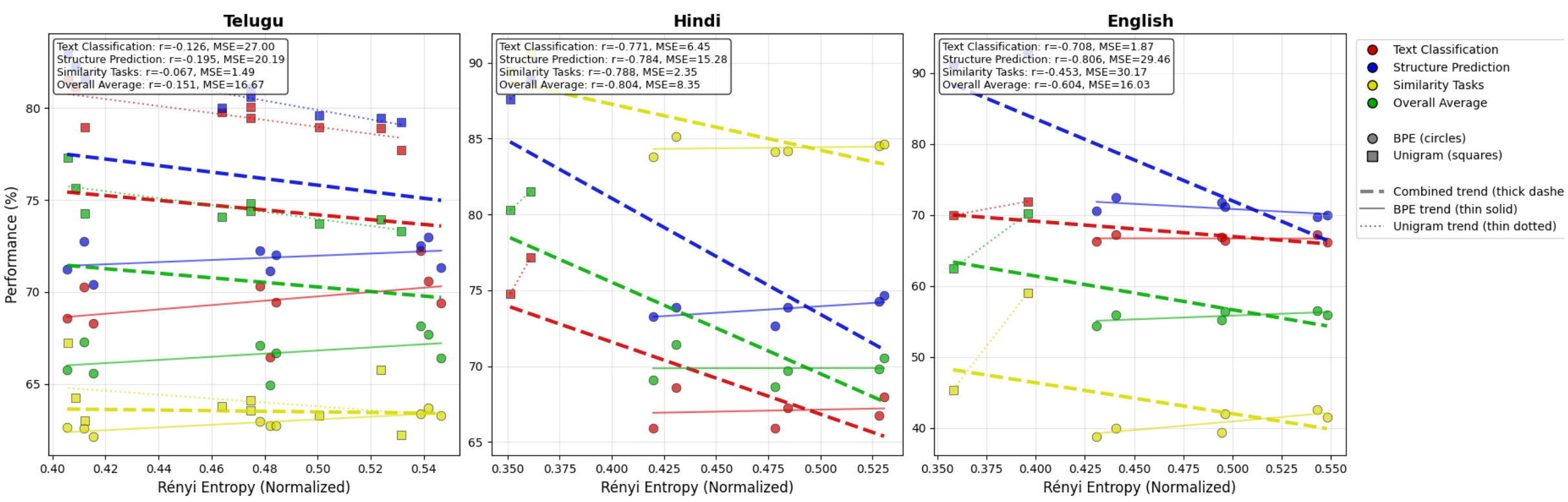

Figure 4: Variation in Downstream Performance with Rényi Entropy (normalized) for different tokenizer variants in Telugu, Hindi, and English.

We also tested the correlation using fixed effects model in order to account for group-level variations. The model included tokenizer and pre-tokenizer as categorical predictors. For structure prediction tasks, both precision (coefficient $= 9.182$, $p = 0.046$) and F1-score (coefficient $= 13.148$, $p = 0.033$) were statistically significant predictors. This implies that after controlling for the choice of tokenizer, a higher morphological precision and F1-score is significantly associated with better performance on structure prediction tasks. For all other tasks, none of the morphological scores show a significant effect once the tokenizer and pre-tokenizer were included in the model.

Since morphological alignment alone cannot account for the large performance differences across all tasks, particularly the consistent success of Unigram tokenizers, we next investigate our second hypothesis, i.e., whether tokenization quality explains the observed trends.

## 5 Tokenization Quality

Another explanation for the observed trend can be that certain tokenizers are inherently more efficient at compressing large data, or they have more efficient distribution of token frequency which helps in better modeling by language model's architecture. We measure compression efficiency using Corpus Token Count (CTC) (Schmidt et al., 2024) and evaluate token frequency distribution using Rényi Entropy (Zouhar et al., 2023).

### 5.1 Corpus Token Count (CTC)

Corpus Token Count (CTC) (Schmidt et al., 2024) is defined as the number of tokens required to encode a given text. It has been argued that better compression leads to improved performance (Gallé, 2019; Goldman et al., 2024). Intuitively, if a tokenizer can represent a text using fewer tokens, it suggests more efficient compression. Thus, a lower CTC is often assumed to indicate better compression and, by extension, better downstream performance. However, our analysis in Telugu, Hindi and English shows that this is not the case for small-sized BERT models at 8.5M scale. Figure 3 shows a plot showing variation in performance with varying CTC. This finding supports previous conclusions by Schmidt et al. (2024); Ali et al. (2024), indicating that compression measured using CTC does not account for the observed variations in downstream task performance across different tokenization settings. We find no statistically significant correlation between CTC and performance on any task. Pearson and Spearman correlations between CTC and performance across all tasks were very weak and not statistically significant. For instance, the pearson correlation between CTC and overall average performance was only $r = 0.135$ ($p = 0.594$), indicating no meaningful linear relationship. We also perform analysis using fixed effects model and found that the coefficient for the logarithm of CTC was not statistically significant. This shows that compression efficiency, atleast as measured using CTC, fails to explain our observed trends in section 3.2.

### 5.2 Rényi Entropy

Zouhar et al. (2023) proposed using an information theoretic measure called Rényi entropy to characterize a good tokenization schema and measure tokenization quality. They contend that Rényi efficiency of the unigram distribution, that a tokenization schema produces, to be the principal measure of tokenization quality. This may also explain the observed performance differences in section 3.2.

We evaluate Rényi entropy for each tokenization variant on a subset of 5 million sentences of our pre-training corpora for corresponding language. Figure 4 shows variation in performance with varying Rényi entropy. We set the parameter $\alpha = 2.5$ as this setting has been found to be the most correlated in Zouhar et al. (2023) with performance. However, we found no statistically significant direct correlation between Rényi entropy and the downstream performance for our small-sized models. While the tokenizer type itself has a major impact on performance, Rényi entropy alone fails to explain the observed trends. Initial correlation tests (both Pearson and Spearman) showed very weak and statistically insignificant relationships between Rényi entropy and performance across all tasks. For instance, the Pearson correlations between Rényi entropy and overall average performance was negligible ($r = -0.151$). Similar to that in section 4.3, ANOVA tests revealed that tokenizer (BPE vs. Unigram) itself has significant effect. The fixed effects models also confirmed these findings. The coefficient of Rényi entropy was consistently not statistically significant across the performance of all tasks. For example, in predicting overall trend, the p-value for the Rényi entropy coefficient was 0.661.

## 6 Discussion

Our findings consistently reveal that Unigram-based tokenizers outperform BPE for small-scale encoder-only BERT models. While successfully demonstrating that this advantage is not explained by intrinsic metrics like Corpus Token Count and Rényi Entropy, the precise reasons for Unigram's success remains unclear. Morphologically-informed pre-tokenization significantly boosted the performance of BPE-based tokenizers, but a similar benefit is not observed for Unigram-based tokenizers.

While morphological alignment showed a moderate yet statistically significant correlation with text classification and structure prediction tasks, it did not fully explain the performance variance across all tasks. Taken together, our results suggest that while linguistic alignment can aid performance, particularly in morphologically rich settings, algorithmic design and vocabulary configuration play a significant role. Designing intrinsic metrics that could consistently explain the performance variations is necessary, and it is important to consider different trade-offs such as between statistical efficiency and linguistic alignment while designing such metrics.

## 7 Conclusion

In this work, we conducted a systematic evaluation of tokenization strategies for languages—Telugu, Hindi and English, with a particular emphasis on agglutinative language—Telugu. Our results shows that morphological alignment have positive corre-lation with downstream effectiveness of tokenizers, while also highlighting the need for more comprehensive intrinsic evaluation of tokenizers which account for various trade-offs.

## Limitations

Our experiments were constrained to encoder-only model and specific to models based on BERT architecture. Therefore there is a potential risk in considering our results generalizable to other architectures. Moreover, we limited our models to 8.5 million parameters. It is not conclusive how our results would scale to larger models. We limited our experiments to three languages with varying degree of morphological complexity. Our conclusion might not be generalizable to all morphologically complex languages, especially given large diversity in morphology across languages. Our evaluations were restricted to natural language understanding (NLU) tasks. Tokenization choices can have different effects on generative tasks (e.g., text summarization, machine translation). Replicating our experiments on other tasks and using models with different architecture might produce considerably different results.

## Ethical considerations

This research was conducted with careful consideration of its ethical dimensions. The models were trained on publicly available corpora, and we acknowledge that these datasets may contain biases from their web-based sources. The primary goal of our work is to positively impact the NLP field by providing a foundation for more equitable and effective models for morphologically complex and under-resourced languages. The new gold-standard morphological dataset created for Telugu is intended for linguistic analysis and is free of any personally identifiable information. As our experiments focus on NLU tasks rather than free-form text generation, the risk of producing harmful content is minimal, though we recognize that the models may still reflect biases from the training data.

In line with our commitment to transparent and reproducible research, we will make all created resources—including the Telugu dataset and all scripts for tokenizer and model training—publicly available. We also acknowledge the significant computational and environmental cost of this work, which involved pre-training 72 models and conducting over 2,160 fine-tuning runs on multiple GPUs. This extensive experimentation was a necessary trade-off to ensure the robustness and validity of our findings.

## Acknowledgments

We would like to express our sincere gratitude to Vandan Mujadia at IIIT Hyderabad for his invaluable guidance and support throughout the various phases of this work. We also thank Nagaraju Vuppala and the language experts at Language Technologies Research Centre (LTRC), IIIT Hyderabad, for their assistance in developing the morphological segmentation tool and also validating the datasets curated in this work.

# A Appendix

## A.1 Pre-training Corpus Statistics

Table 4 presents the detailed statistics of the corpus used in this study. TTR represents Type-Token Ratio, MATTR represent Mean-Average Type-Token Ratio and MLW represent Mean Length of Word across the corpus.

| Metrics | English | Hindi | Telugu |
|---|---|---|---|
| #Sentences | 10,000,000 | 10,095,405 | 6,721,543 + 3,278,457* = 10,000,000 |
| #Tokens | 198,637,872 | 169,127,701 | 95,063,928 |
| #Types | 2,499,750 | 1,462,501 | 4,206,880 |
| TTR | 0.012584 | 0.008647 | 0.0442531 |
| MLW | 4.7760 | 4.0329 | 6.8989 |
| MATTR[†] | 0.8051 | 0.4674 | 0.4167 |

Table 4: Corpus statistics of different languages used for training tokenizers and pretraining the language models. * indicates data from the IndicCorp dataset. Metrics marked with † are calculated on 100 million character subset of the corpus.

## A.2 Morphological Alignment

Figures 5, 6 and 7 shows downstream performance vs. morphological alignment trends for Telugu, Hindi, and English respectively. Tables 5, 6 and 7 presents detailed MorphScores for each tokenization variant across Telugu, Hindi and English respectively.

To quantify the degree to which a tokenizer's segmentations align with linguistic morpheme boundaries, we employ the **MorphScore** evaluation metric (Arnett and Bergen, 2024; Arnett et al., 2025). This is a boundary-based method that compares the segmentation points produced by a tokenizer against a gold standard set of morpheme boundaries for a given list of words. The evaluation proceeds as follows for each word in the test set:

1. **Boundary Identification:** Both the gold-standard morphemic segmentation and the tokenizer's output are converted into sets of character-level boundary indices. For a word of length $N$, a boundary is an integer index $i$ (from $1$ to $N-1$) that marks the end of a segment. This results in a gold set, $B_{gold}$, and a predicted set, $B_{pred}$.

2. **Metric Calculation:** Using these sets, we calculate True Positives (TP), False Positives (FP), and False Negatives (FN) to assess the alignment:

   - **True Positives (TP):** The number of boundaries correctly identified by the tokenizer. This corresponds to the size of the intersection of the two sets.

   $$\text{TP} = |B_{gold} \cap B_{pred}|$$

   - **False Positives (FP):** The number of boundaries predicted by the tokenizer that do not exist in the gold standard. This indicates over-segmentation.

   $$\text{FP} = |B_{pred} - B_{gold}|$$

   - **False Negatives (FN):** The number of gold-standard boundaries missed by the tokenizer. This indicates under-segmentation.

   $$\text{FN} = |B_{gold} - B_{pred}|$$

   From these counts, we compute **Precision**, **Recall**, and the **F1-score** for each word:

   $$\text{Precision} = \frac{\text{TP}}{\text{TP} + \text{FP}}$$

   $$\text{Recall} = \frac{\text{TP}}{\text{TP} + \text{FN}}$$

   $$\text{F1} = 2 \cdot \frac{\text{Precision} * \text{Recall}}{\text{Precision} + \text{Recall}}$$

3. **Exclusion Criteria:** To ensure a fair evaluation focused on complex word forms, certain words are excluded from the calculation:

   - Words that the tokenizer outputs as a single token (i.e., where $B_{pred}$ is empty).
   - Words that consist of a single morpheme in the gold standard (i.e., where $B_{gold}$ is empty).

4. **Final Score Aggregation:** The final Recall, Precision, and F1-score for a given tokenizer are the micro-averages of these metrics calculated over all non-excluded words in the evaluation dataset.

**Example Calculation**

Consider the Telugu word ఆధారపడతాము (length 11).

- **Gold Segmentation:** ఆధారపడ + త + ము
  The morpheme boundaries are after the 6th character ('డ') and the 8th character ('త').

$$B_{gold} = \{6, 8\}$$

- **Tokenizer Output (Morfessor + BPE):** ఆధారపడ + తాము
  The tokenizer places one boundary after the 6th character ('డ').

$$B_{pred} = \{6\}$$

The metrics are then calculated as follows:

- $\text{TP} = |\{6, 8\} \cap \{6\}| = 1$
- $\text{FP} = |\{6\} - \{6, 8\}| = 0$
- $\text{FN} = |\{6, 8\} - \{6\}| = 1$

$$\text{Recall} = \frac{1}{1+1} = 0.5$$

$$\text{Precision} = \frac{1}{1+0} = 1.0$$

This indicates that while every boundary the tokenizer predicted was correct (high precision), it only found half of the true morpheme boundaries (lower recall).

## A.3 Downstream Tasks Description

To assess the performance across tokenization variants, we utilize an extensive set of downstream tasks, verified and suitable for each language. These tasks span diverse categories including Classification, Structure Prediction, Question Answering, and Natural Language Inference. We provide details below, organized by language, with overlapping tasks clearly indicated.

### A.3.1 Tasks Description

**Word & Definition** (WaD): Classify whether a given word and a given definition match semantically (Batsuren et al., 2024).

| word | definition | label |
|---|---|---|
| clerking | the activity of recording business transactions | 1 |
| ammo | alternatively placed in genus Martynia | 0 |
| enforced | forced or compelled or put in force | 1 |
| snowline | a fishing line managed principally by hand | 0 |

Table 8: Example data points in Word and Definition task.

**Word & Morphology** (WaM): Classify whether a given word contains inflection, derivation, or compounding (Batsuren et al., 2024).

| word | morphology | label |
|---|---|---|
| leaderboard | derivation | 1 |
| overpressing | compound | 0 |
| coteaches | inflection | 1 |
| sharemarkets | derivation | 0 |

Table 9: Example data points in Word and Morphology task.

**Word & Word** (WaW): Classify whether two given words are semantically related (Batsuren et al., 2024). For Telugu and Hindi, we utilize IndicWordNet[3] (Kanojia et al., 2022), accessing through API[4] (Panjwani et al., 2018). We follow similar steps as mentioned in Batsuren et al. (2024) while curating the data. The resulting dataset is further manually validated by language experts to ensure correctness.

For each synset in IndicWordNet, we extract the *head word* and collect words connected through semantic relations such as SIMILAR, HYPERNYMY, and HYPONYMY. Word pairs are then formed between the head word and each related word. Pairs containing special characters or identical words are discarded, and duplicates are removed. The resulting pairs are assigned the label **1 (related)**.

Negative pairs are generated to ensure semantic unrelatedness. For each positive pair, a random candidate pair is sampled from the vocabulary, subject to strict constraints: (i) the two words must not be identical, (ii) the pair or its reverse must not exist among positive samples, (iii) the words must not share neighbors in the semantic graph, (iv) the words must not share hypernyms, and (v) the words must not be connected by entailment. A maximum attempt limit is enforced to prevent infinite loops. All validated pairs are labeled **0 (unrelated)**.

The curated dataset is stored in TSV format with columns: `index`, `word_a`, `word_b`, and `label`. Finally, the dataset is manually validated by language experts to ensure correctness. The dataset can be found here: IndicSigmorphon-Dataset.

**Parts of Speech Tagging** (POS): Assigning grammatical category (such as noun, verb, adjective, etc.) to each word in a sentence based on both its definition and its context within the sentence (Nivre et al., 2020). For deciding the class of a word given subword classes, we report both results considering first token class and max-pooling of the classes of each token. We refer first token class based classification as POS, while max-pooling as POS-Pooled in all our results.

[3] https://www.cfilt.iitb.ac.in/indowordnet/
[4] https://github.com/cfiltnlp/pyiwn

| Pre-tokenizer | Tokenizer | Vocabulary Size | Recall | Precision | F1-score |
|---|---|---|---|---|---|
| - | Character | - | 1.0000 | 0.145931 | 0.254695 |
| - | BPE | 8192 | 0.3118 | 0.3118 | 0.313208 |
| | BPE | 16384 | 0.2789 | 0.2527 | 0.265214 |
| | BPE | 50277 | 0.2661 | 0.2707 | 0.268367 |
| Morfessor | BPE | 8192 | 0.3406 | 0.2599 | 0.294857 |
| | BPE | 16384 | 0.3111 | 0.2896 | 0.300285 |
| | BPE | 50277 | 0.2785 | 0.3053 | 0.291013 |
| Morph Analyzer | BPE | 8192 | 0.6257 | 0.4463 | 0.521033 |
| | BPE | 16384 | 0.5757 | 0.4983 | 0.534195 |
| | BPE | 50277 | 0.5190 | 0.5291 | 0.523997 |
| - | Unigram | 8192 | 0.3924 | 0.3544 | 0.372837 |
| | Unigram | 16384 | 0.3950 | 0.3572 | 0.375818 |
| | Unigram | 50277 | 0.4146 | 0.4211 | 0.417517 |
| Morfessor | Unigram | 8192 | 0.3852 | 0.2526 | 0.305000 |
| | Unigram | 16384 | 0.4079 | 0.3307 | 0.364393 |
| | Unigram | 50277 | 0.3949 | 0.3674 | 0.380777 |
| Morph Analyzer | Unigram | 8192 | 0.7774 | 0.5505 | 0.644564 |
| | Unigram | 16384 | 0.7385 | 0.6368 | 0.683873 |
| | Unigram | 50277 | 0.8046 | 0.7769 | 0.790517 |

Table 5: MorphScores of various tokenization strategies using different tokenizer variants across various vocabulary sizes in Telugu.

**ACTSA** (Annotated Corpus for Telugu Sentiment Analysis): Determine the sentiment associated with a sentence (sentiment analysis) (Mukku and Mamidi, 2017). This task is specifically curated for Telugu by native Telugu speakers.

**IndicSentiment**: Sentiment analysis on synthetically created product reviews introduced in Doddapaneni et al. (2023). This task presents a 13-way parallel dataset, with sentences synthetically created for English and later translated to Indian languages. This dataset claims to avoid one-dimensional and highly polarized product reviews (makes classification easier).

**IIT-Patna Movie Reviews & Product Reviews**: Includes sentiment analysis task with dataset specifically curated by using reviews posted in Hindi (Akhtar et al., 2016). These datasets has 4 classes namely positive, negative, neutral, and conflict.

**MASSIVE Intent Classification**: Multilingual Amazon Slu resource package for Intent Classification. This dataset was introduced in FitzGerald et al. (2023) and was created using user queries collected by Amazon Alexa. The dataset contains 60 intents.

**Named Entity Recognition** (NER): Involves identifying and classifying named entities in text into predefined categories such as persons, organizations, locations, dates and other proper nouns. For Telugu and Hindi, we use dataset from WikiAnn[5] (Pan et al., 2017; Doddapaneni et al., 2023). The dataset consists of coarse grained labels: Person (PER), Organization (ORG) and Location (LOC). While for English, we use CoNLL NER dataset (Sang and Meulder, 2003). It contains predefined categories such as Person (PER), Organization (ORG), Location (LOC), and Miscellaneous (MISC).

**IndicXParaphrase**: This task involves classifying whether a pair of sentences are paraphrased or not (Kumar et al., 2022; Doddapaneni et al., 2023)[6]. Each entry in the dataset is a tuple

[5] https://elisa-ie.github.io/wikiann/

[6] https://huggingface.co/datasets/ai4bharat/IndicXParaphrase

| Pre-tokenizer | Tokenizer | Vocabulary Size | Recall | Precision | F1-score |
|---|---|---|---|---|---|
| - | Character | - | 1.0000 | 0.1414 | 0.247791 |
| - | BPE | 8192 | 0.7312 | 0.1484 | 0.246745 |
| | BPE | 16384 | 0.6659 | 0.1538 | 0.249938 |
| | BPE | 50277 | 0.5247 | 0.1519 | 0.235697 |
| Morfessor | BPE | 8192 | 0.7667 | 0.1571 | 0.257695 |
| | BPE | 16384 | 0.6824 | 0.1653 | 0.266085 |
| | BPE | 50277 | 0.4862 | 0.1707 | 0.252684 |
| - | Unigram | 8192 | 0.8099 | 0.1748 | 0.287547 |
| | Unigram | 16384 | 0.7657 | 0.1929 | 0.301085 |
| | Unigram | 50277 | 0.6759 | 0.2286 | 0.341597 |
| Morfessor | Unigram | 8192 | 0.8155 | 0.1902 | 0.308457 |
| | Unigram | 16384 | 0.6549 | 0.2026 | 0.309463 |
| | Unigram | 50277 | 0.5237 | 0.1907 | 0.279579 |

Table 6: MorphScores of various tokenization strategies using different tokenizer variants across various vocabulary sizes in Hindi.

| Pre-tokenizer | Tokenizer | Vocabulary Size | Recall | Precision | F1-score |
|---|---|---|---|---|---|
| - | Character | - | 1.0000 | 0.1414 | 0.247791 |
| - | BPE | 8192 | 0.3049 | 0.1299 | 0.18226 |
| | BPE | 16384 | 0.2483 | 0.1105 | 0.15295 |
| | BPE | 50277 | 0.2163 | 0.0975 | 0.134405 |
| Morfessor | BPE | 8192 | 0.5238 | 0.2241 | 0.313355 |
| | BPE | 16384 | 0.5078 | 0.2356 | 0.321848 |
| | BPE | 50277 | 0.4916 | 0.2378 | 0.32052 |
| - | Unigram | 8192 | 0.8929 | 0.3351 | 0.487372 |
| | Unigram | 16384 | 0.8515 | 0.3541 | 0.500189 |
| | Unigram | 50277 | 0.8146 | 0.3732 | 0.517995 |
| Morfessor | Unigram | 8192 | 0.9209 | 0.3115 | 0.465555 |
| | Unigram | 16384 | 0.9111 | 0.3222 | 0.476074 |
| | Unigram | 50277 | 0.9063 | 0.3229 | 0.476204 |

Table 7: MorphScores of various tokenization strategies using different tokenizer variants across various vocabulary sizes in English.

| word | word | label |
|---|---|---|
| visitor | traveler | 1 |
| shopper | earless | 0 |
| photocopy | mosaic | 1 |
| bleed | medicine | 1 |

Table 10: Example datapoints in Word and Word task.

<English_sentence, sentence-1, sentence-2>, where sentence-1 and sentence-2 refer to pairs of sentences.

**Natural Language Inference** (NLI): Includes multilingual natural language inference benchmark that evaluates a model's ability to determine the logical relationship-entailment, contradictions, or neutrality-between pairs of sentences, called premise and hypothesis. (Aggarwal et al., 2022; Conneau et al., 2018).

**Discourse Mode Classification** (DM): Identifying the discourse mode or textual function of a given Hindi sentence or paragraph. A discourse mode represents the communicative purpose or rhetorical function of a segment of text. The dataset contains five different discourse modes: *argumentative, narrative, descriptive, dialogic, and informative* (Dhanwal et al., 2020).

**STS-B**: The Semantic Textual Similarity task (Cer et al., 2017) is a collection of sentence pairs drawn from news headlines, video and image captions, and natural language inference data. Each pair is

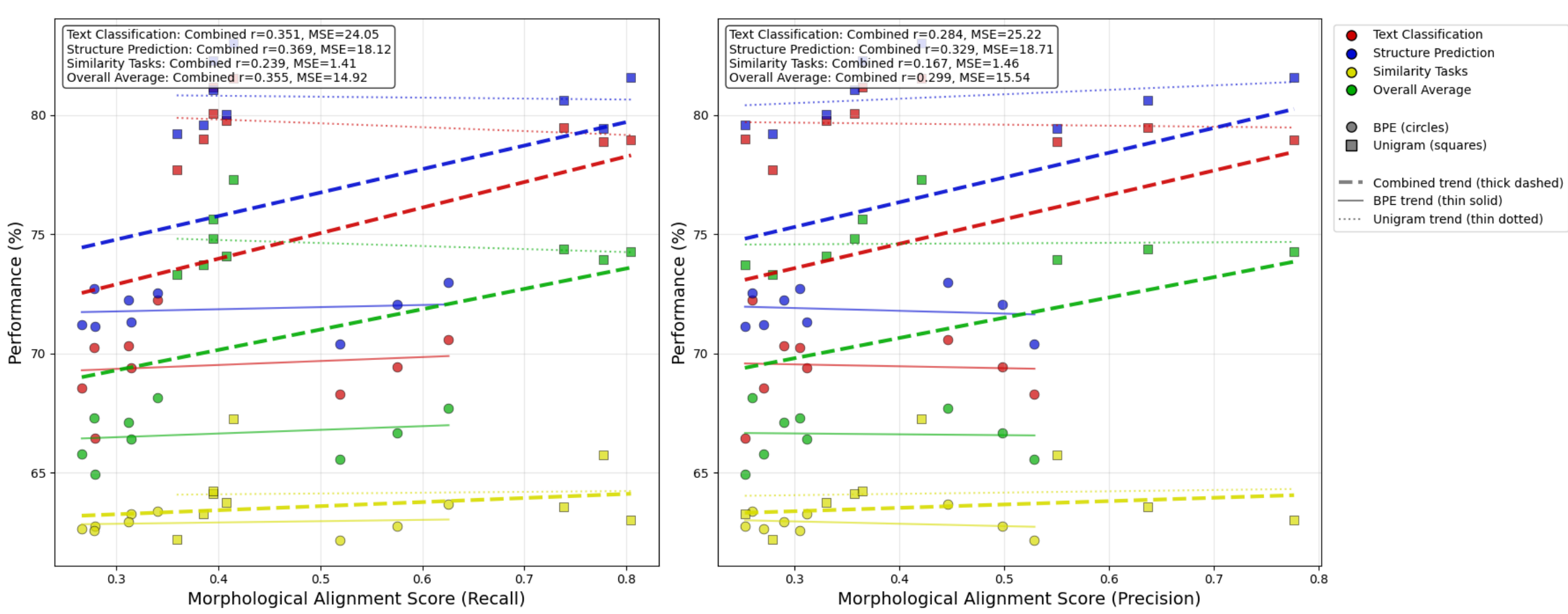

Figure 5: Variation in Downstream Performance with Morphological Alignment Scores for different tokenizer variants in Telugu.

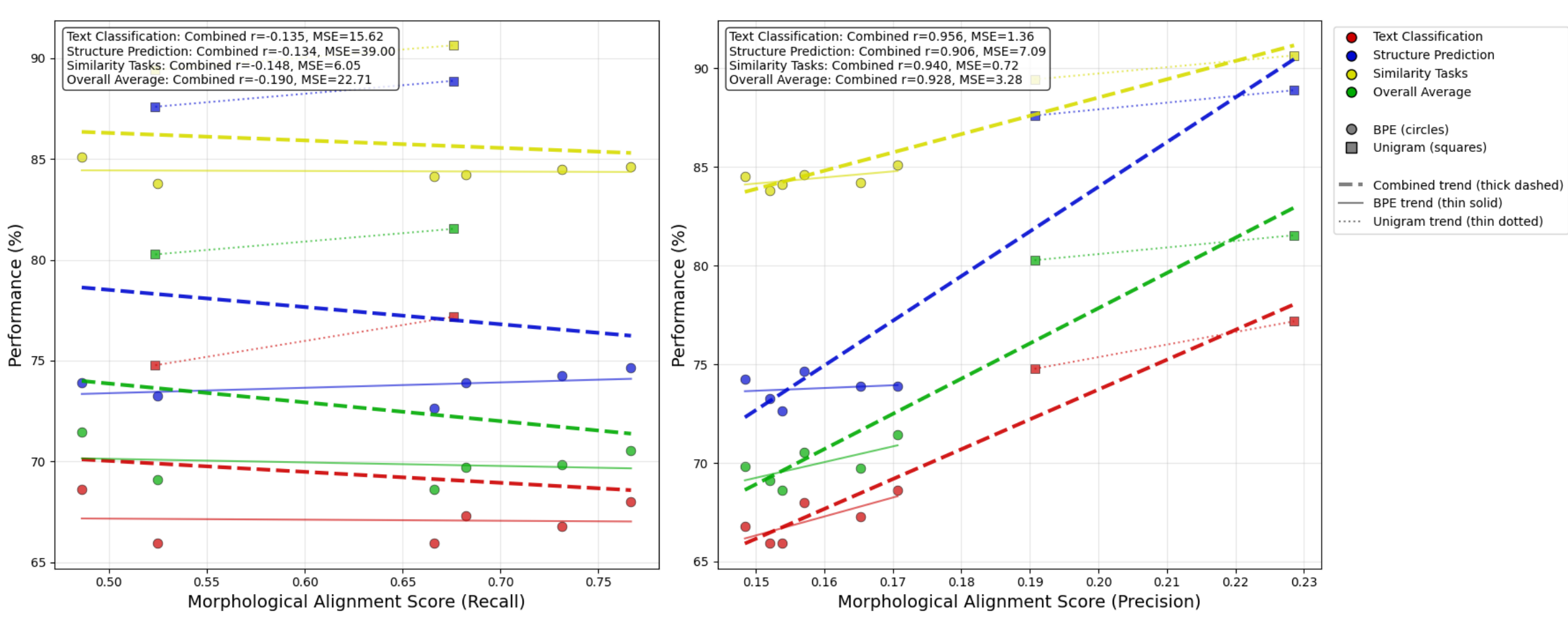

Figure 6: Variation in Downstream Performance with Morphological Alignment Scores for different tokenizer variants in Hindi.

human-annotated with a similarity score from 1 to 5. The task involves predicting these scores. Evaluation metrics includes Pearson and Spearman correlation coefficients. (Wang et al., 2019)

**Dependency Parsing**: Involved analyzing the grammatical structure of a sentence by identifying relationships between "head" words and their dependents. We used Universal Dependencies (UD) Treebank dataset (Nivre et al., 2020) to perform dependency parsing. The model was adapted to predict both the syntactic head of each word and the type of dependency relation. Performance was evaluated using standard metrics: Unlabeled Attachment Score (UAS) and Labeled Attachment Score (LAS) (Nivre and Fang, 2017).

### A.4 Pre-training Hyperparameters

Hyperparameters settings of BERT models in our experiments are shown in Table 11. Each model contained approximately 8.5 million parameters excluding the parameters in embedding layer.

### A.5 Fine-tuning Hyperparameters

We adopt hyperparameter settings from prior work, as our experiments focus solely on comparative evaluation. Consequently, we did not find it necessary to perform additional hyperparameter tuning. Details regarding specific hyperparameter for each task can be found in table 12.

### A.6 Downstream Performance

We evaluated performance of languages models on extensive set of downstream tasks ranging from Sequence Classification, Parts-of-Speech Tagging to Natural Language Inference Tasks such as IndicXNLI. Tables 13, 14, and 15 show performance across various languages and downstream tasks.

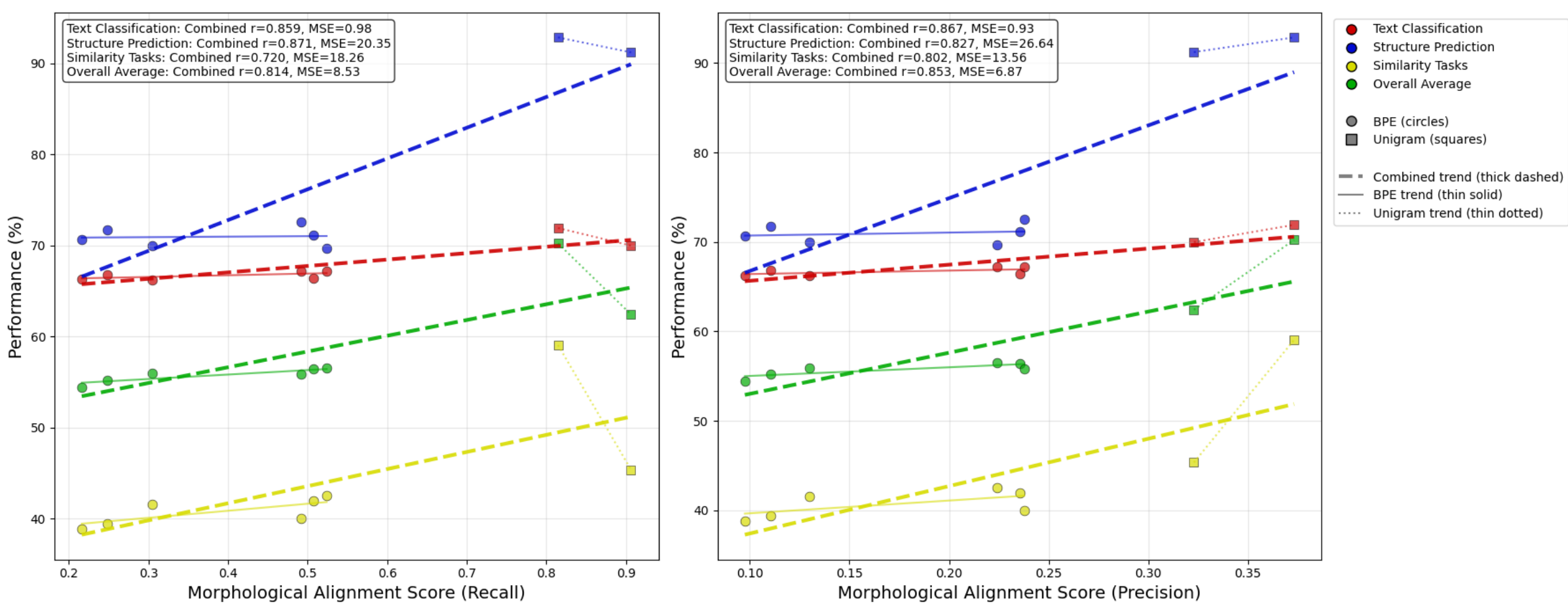

Figure 7: Variation in Downstream Performance with Morphological Alignment Scores for different tokenizer variants in English.

| Hyperparameter | Value |
|---|---|
| Batch size | 128 |
| Total training steps | 175,000 |
| Adam $\epsilon$ | $1 \times 10^{-6}$ |
| Adam $\beta_1$ | 0.9 |
| Adam $\beta_2$ | 0.999 |
| Sequence length | 128 |
| Learning rate | $1 \times 10^{-4}$ |
| Learning rate schedule | Linear warmup |
| Warmup steps | 3,750 |
| Weight decay | 0.01 |
| Attention dropout | 0.1 |
| Dropout | 0.1 |
| Hidden Size | 384 |
| Number of Attention Heads | 6 |
| Number of Hidden Layers | 6 |

Table 11: Hyperparameters choices of the BERT language models pre-trained for our evaluations.

| Hyperparameter | Value |
|---|---|
| Train Batch size | 16 for POS, IndicXNLI, STS-B, Dependency Parsing<br>32 for WaD, WaM, WaW, ACTSA, IndicSentiment, IITP-MR, IITP-PR, MASSIVE, IndicX-Para, DM |
| Eval Batch size | 16 for POS, IndicXNLI, STS-B, Dependency Parsing<br>32 for WaD, WaM, WaW, ACTSA, IndicSentiment, IITP-MR, IITP-PR, Wiki-NER, IndicX-Para, DM<br>64 for MASSIVE |
| Epochs | 5 for WaD, WaM, WaW, POS, IndicXNLI<br>10 for ACTSA, IndicSentiment, IITP-MR, IITP-PR, MASSIVE, Wiki-NER, DM, STS-B<br>20 for IndicXPara |
| Adam $\epsilon$ | 1e-8 |
| Adam $\beta_1$ | 0.9 |
| Adam $\beta_2$ | 0.999 |
| Learning rate | 2e-5 for ACTSA, IndicSentiment, IITP-MR, IITP-PR, IndicXPara, DM<br>3e-5 for WaD, WaM, WaW, IndicXNLI, STS-B<br>5e-5 for MASSIVE, Wiki-NER |
| Learning rate schedule | Linear warmup |
| Warmup steps | 10% of steps |
| Weight decay | 0.01 |
| Attention dropout | 0.1 |
| Dropout | 0.1 |
| Sequence Length | 128 |

Table 12: Hyperparameters used for fine-tuning for each downstream task.

| Pre-tokenizer | Tokenizer | Vocab Size | Downstream Tasks | | | | |
|---|---|---|---|---|---|---|---|
| | | | WaW<br>Accuracy | POS<br>F1 | POS-Pooled<br>F1 | ACTSA<br>Accuracy | IndicSentiment<br>Accuracy |
| None (Naive) | Character | - | 78.54±0.60 | 83.63±0.11 | 84.44±0.10 | 55.41±1.55 | 69.23±2.93 |
| | | 8192 | 78.92±0.58 | 78.72±0.11 | 86.18±0.04 | 53.61±0.00 | 72.44±0.00 |
| None (Naive) | BPE | 16384 | 77.86±0.47 | 77.52±0.06 | 85.37±0.08 | 54.16±0.00 | 62.18±0.00 |
| | | 50277 | 77.64±0.31 | 78.24±0.36 | 85.86±0.16 | 53.23±0.00 | 72.44±0.00 |
| | | 8192 | 79.15±0.39 | 79.94±0.18 | 86.40±0.26 | 56.26±0.00 | 78.85±0.00 |
| Morfessor | BPE | 16384 | 78.28±0.20 | 78.38±0.28 | 85.74±0.02 | 54.16±0.00 | 76.28±0.00 |
| | | 50277 | 77.51±0.17 | 79.43±0.58 | 86.26±0.25 | 54.53±0.00 | 75.64±0.00 |
| | | 8192 | 79.72±0.21 | 80.69±0.06 | 86.28±0.20 | 56.38±0.00 | 73.72±0.00 |
| Morph Analyzer | BPE | 16384 | 77.88±0.26 | 79.27±0.25 | 85.49±0.18 | 53.97±0.00 | 73.72±0.00 |
| | | 50277 | 76.66±0.18 | 77.87±0.17 | 84.96±0.05 | 56.01±0.00 | 69.87±0.00 |
| | | 8192 | 84.53±0.24 | 83.62±0.28 | 86.16±0.11 | 64.51±0.00 | 85.26±0.00 |
| None (Naive) | Unigram | 16384 | 86.33±0.25 | 84.20±0.28 | 85.32±0.00 | 67.10±0.00 | 85.26±0.00 |
| | | 50277 | 89.36±0.13 | 87.76±0.22 | 86.99±0.20 | 66.54±0.00 | 87.18±0.00 |
| | | 8192 | 84.91±0.32 | 84.07±0.14 | 86.28±0.17 | 65.80±-0.00 | 84.62±0.00 |
| Morfessor | Unigram | 16384 | 86.90±0.15 | 85.02±0.10 | 85.55±0.00 | 67.47±0.00 | 82.69±0.00 |
| | | 50277 | 88.82±0.09 | 88.71±0.18 | 86.95±0.55 | 65.80±0.00 | 87.18±0.00 |
| | | 8192 | 84.87±0.15 | 81.73±0.15 | 86.21±0.05 | 64.51±0.00 | 85.26±0.00 |
| Morph Analyzer | Unigram | 16384 | 77.88±0.26 | 80.00±0.20 | 85.49±0.18 | 53.97±0.00 | 73.72±0.00 |
| | | 50277 | 87.63±0.06 | 83.52±0.10 | 85.64±0.20 | 64.14±0.00 | 82.69±0.00 |
| | | 8192 | 72.19±0.07 | 34.22±0.11 | 34.66±0.29 | 58.23±0.00 | 75.00±0.00 |
| Morph Analyzer | Word | 16384 | 74.07±0.06 | 34.17±0.20 | 34.92±0.12 | 57.86±0.00 | 75.64±0.00 |
| | | 50277 | 76.53±0.19 | 34.57±0.04 | 57.90±0.03 | 58.60±0.00 | 75.64±0.00 |
| | | 8192 | 70.98±0.11 | 44.14±0.05 | 35.04±0.24 | 63.50±0.00 | 78.20±0.00 |
| Morfessor | Word | 16384 | 75.19±0.14 | 59.46±0.07 | 34.88±0.08 | 62.48±0.00 | 76.28±0.00 |
| | | 50277 | 79.45±0.09 | 72.50±0.03 | 59.46±0.06 | 63.96±0.00 | 82.05±0.00 |
| | | 8192 | 70.36±0.14 | 38.82±0.04 | 34.97±0.21 | 65.06±0.00 | 76.28±0.00 |
| None (Naive) | Word | 16384 | 74.98±0.09 | 55.17±0.06 | 34.95±0.23 | 60.63±0.00 | 79.49±0.00 |
| | | 50277 | 77.97±0.08 | 64.16±0.02 | 54.46±0.09 | 63.77±0.00 | 82.05±0.00 |

| Pre-tokenizer | Tokenizer | Vocab Size | Downstream Tasks | | | | |
|---|---|---|---|---|---|---|---|
| | | | Massive Intent<br>Accuracy | Wiki-NER<br>F1 | IndicXPara<br>Accuracy | IndicXNLI<br>F1 | Dependency Parsing<br>LAS/UAS |
| None (Naive) | Character | - | 75.42±0.71 | 88.75±0.05 | 44.06±3.25 | 51.45±1.03 | 50.82±1.03/62.09±1.72 |
| | | 8192 | 73.00±0.00 | 87.97±0.29 | 47.63±0.00 | 53.15±0.47 | 48.82±0.47/62.34±0.53 |
| None (Naive) | BPE | 16384 | 71.57±0.00 | 86.72±0.23 | 47.63±0.00 | 51.59±0.16 | 49.84±0.94/62.60±0.99 |
| | | 50277 | 70.98±0.00 | 86.69±0.10 | 47.63±0.00 | 51.03±0.56 | 49.74±0.32/62.60±0.99 |
| | | 8192 | 74.96±0.00 | 87.96±0.09 | 47.63±0.00 | 54.62±0.06 | 51.75±0.73/63.99±0.89 |
| Morfessor | BPE | 16384 | 72.50±0.00 | 87.67±0.08 | 47.63±0.00 | 53.20±0.36 | 50.87±0.79/64.76±0.44 |
| | | 50277 | 73.34±0.00 | 88.00±0.16 | 47.63±0.00 | 53.24±0.34 | 51.29±0.58/65.43±0.46 |
| | | 8192 | 73.09±0.00 | 88.59±0.25 | 47.63±0.00 | 54.59±0.48 | 51.75±0.62/65.28±0.86 |
| Morph Analyzer | BPE | 16384 | 72.11±0.00 | 87.12±0.12 | 47.63±0.00 | 53.34±0.40 | 50.82±0.73/64.71±1.25 |
| | | 50277 | 70.68±0.00 | 86.33±0.16 | 47.63±0.00 | 53.38±0.38 | 48.51±0.70/61.78±0.76 |
| | | 8192 | 81.36±0.00 | 94.01±0.16 | 39.90±0.00 | 60.68±0.28 | 63.39±0.30/73.37±1.09 |
| None (Naive) | Unigram | 16384 | 81.55±0.00 | 94.58±0.06 | 41.90±0.00 | 61.90±0.06 | 67.07±1.23/77.32±1.18 |
| | | 50277 | 83.18±0.00 | 95.78±0.04 | 45.14±0.00 | 69.51±0.22 | 70.46±1.15/78.81±1.23 |
| | | 8192 | 80.96±0.00 | 94.16±0.14 | 41.65±0.00 | 60.93±0.06 | 63.99±0.76/73.95±0.61 |
| Morfessor | Unigram | 16384 | 82.05±0.00 | 94.42±0.06 | 40.65±0.00 | 61.08±0.04 | 65.29±0.35/74.86±0.75 |
| | | 50277 | 82.83±0.00 | 95.50±0.14 | 39.65±0.00 | 62.98±0.68 | 68.89±0.40/77.74±0.38 |
| | | 8192 | 80.92±0.00 | 94.42±0.08 | 46.63±0.00 | 59.31±0.12 | 63.78±0.78/73.40±0.40 |
| Morph Analyzer | Unigram | 16384 | 72.11±0.00 | 94.91±0.15 | 41.15±0.00 | 62.36±0.15 | 66.16±1.30/75.66±1.02 |
| | | 50277 | 81.36±0.00 | 95.14±0.12 | 38.40±0.00 | 62.15±0.28 | 67.94±1.55/77.59±1.59 |
| | | 8192 | 57.65±0.00 | 90.59±0.18 | 43.14±0.00 | 52.90±0.03 | 69.52±1.20/84.79±0.47 |
| Morph Analyzer | Word | 16384 | 60.45±0.00 | 91.78±0.09 | 38.65±0.00 | 52.88±0.02 | 69.12±0.32/83.56±0.09 |
| | | 50277 | 65.72±0.00 | 93.27±0.13 | 43.64±0.00 | 54.18±0.13 | 69.06±0.55/83.41±0.69 |
| | | 8192 | 60.85±0.00 | 91.21±0.06 | 41.15±0.00 | 55.54±0.34 | 71.36±0.80/84.74±0.62 |
| Morfessor | Word | 16384 | 64.24±0.00 | 92.27±0.03 | 41.40±0.00 | 55.62±0.29 | 73.81±0.55/85.91±0.85 |
| | | 50277 | 68.47±0.00 | 93.84±0.18 | 35.91±0.00 ‘ | 57.73±0.50 | 74.94±0.69/86.12±0.35 |
| | | 8192 | 63.26±0.00 | 91.75±0.08 | 41.65±0.00 | 57.73±0.06 | 73.81±0.55/85.55±0.64 |
| None (Naive) | Word | 16384 | 65.76±0.00 | 92.87±0.076 | 40.15±0.00 | 57.03±0.60 | 73.81±1.53/84.43±0.87 |
| | | 50277 | 72.21±0.00 | 94.43±0.051 | 37.66±0.00 | 61.86±0.03 | 74.27±0.67/85.30±0.26 |

Table 13: Comparison of downstream performances across tokenizers in Telugu.

| Pre-tokenizer | Tokenizer | Vocab Size | Downstream Tasks | | | | |
|---|---|---|---|---|---|---|---|
| | | | **WaW** Accuracy | **POS** F1 | **POS-Pooled** F1 | **IITP-MR** Accuracy | **IITP-PR** Accuracy |
| None (Naive) | Character | - | 85.36±0.28 | 83.35±0.21 | 79.32±0.87 | 47.96±0.19 | 67.11±0.33 |
| | | 8192 | 84.50±0.10 | 81.01±0.27 | 83.71±0.27 | 47.10±0.00 | 61.76±0.00 |
| None (Naive) | BPE | 16384 | 84.13±0.71 | 80.03±0.21 | 82.58±0.14 | 43.55±0.00 | 62.14±0.00 |
| | | 50277 | 83.81±0.34 | 81.28±0.53 | 82.35±0.31 | 44.52±0.00 | 62.14±0.00 |
| | | 8192 | 84.62±0.13 | 80.68±0.50 | 83.93±0.10 | 48.39±0.00 | 64.63±0.00 |
| Morfessor | BPE | 16384 | 84.21±0.01 | 80.88±0.09 | 83.20±0.11 | 46.45±0.00 | 64.44±0.00 |
| | | 50277 | 85.11±0.07 | 82.24±0.26 | 84.40±0.24 | 49.03±0.00 | 64.82±0.00 |
| | | 8192 | -±- | -±- | -±- | -±- | -±- |
| None (Naive) | Unigram | 16384 | -±- | -±- | -±- | -±- | -±- |
| | | 50277 | 90.64±0.06 | 89.50±0.09 | 96.82±0.00 | 62.26±0.00 | 77.06±0.00 |
| | | 8192 | -±- | -±- | -±- | -±- | -±- |
| Morfessor | Unigram | 16384 | -±- | -±- | -±- | -±- | -±- |
| | | 50277 | 89.43±0.20 | 89.80±0.05 | 95.67±0.05 | 57.42±0.00 | 74.19±0.00 |
| | | 8192 | 78.84±0.02 | 48.19±0.11 | 94.21±0.12 | 59.36±0.00 | 73.04±0.00 |
| Morfessor | Word | 16384 | 82.73±0.05 | 64.20±0.12 | 92.65±0.11 | 59.68±0.00 | 73.23±0.00 |
| | | 50277 | -±- | -±- | -±- | -±- | -±- |
| | | 8192 | 77.87±0.03 | 38.38±0.06 | 95.42±0.03 | 57.10±0.00 | 74.00±0.00 |
| None (Naive) | Word | 16384 | 83.44±0.05 | 54.53±0.07 | 93.83±0.04 | 61.29±0.00 | 76.86±0.00 |
| | | 50277 | 87.25±0.03 | 71.12±0.02 | 97.16±0.02 | 60.64±0.00 | 78.78±0.00 |

| Pre-tokenizer | Tokenizer | Vocab Size | Downstream Tasks | | | | |
|---|---|---|---|---|---|---|---|
| | | | **DM** Accuracy | **Wiki-NER** F1 | **IndicXPara** Accuracy | **IndicXNLI** F1 | **Dependency Parsing** LAS/UAS |
| None (Naive) | Character | - | 74.36±0.12 | 86.60±0.28 | 47.13±1.95 | 51.49±- | 58.83±0.74/66.71±0.62 |
| | | 8192 | 73.72±0.00 | 86.08±0.29 | 62.84±0.00 | 53.26±- | 59.44±0.33/67.74±0.31 |
| None (Naive) | BPE | 16384 | 73.92±0.00 | 84.13±0.42 | 46.88±0.00 | 51.87±- | 57.86±0.31/65.99±0.36 |
| | | 50277 | 73.22±0.00 | 84.59±0.62 | 63.84±0.00 | 52.86±- | 58.89±0.29/67.18±0.23 |
| | | 8192 | 74.32±0.00 | 86.20±0.40 | 61.84±0.00 | 53.93±- | 60.18±0.08/68.28±0.10 |
| Morfessor | BPE | 16384 | 74.02±0.00 | 85.50±0.64 | 62.34±0.00 | 51.65±- | 59.38±0.39/67.51±0.37 |
| | | 50277 | 75.43±0.00 | 86.66±0.21 | 61.10±0.00 | 53.77±- | 62.59±0.13/70.43±0.14 |
| | | 8192 | -±- | -±- | -±- | -±- | -±-/-±- |
| None (Naive) | Unigram | 16384 | -±- | -±- | -±- | -±- | -±-/-±- |
| | | 50277 | 90.64±0.06 | 83.22±0.21 | -±- | 61.64±- | 85.83±0.08/89.66±0.02 |
| | | 8192 | -±- | -±- | -±- | -±- | -±-/-±- |
| Morfessor | Unigram | 16384 | -±- | -±- | -±- | -±- | -±-/-±- |
| | | 50277 | 89.43±0.20 | 83.65±0.24 | -±- | 63.44±- | 83.37±0.05/87.69±0.04 |
| | | 8192 | 77.53±0.00 | 87.37±0.22 | 81.30±0.00 | 60.61±- | 83.54±0.07/88.67±0.05 |
| Morfessor | Word | 16384 | 76.93±0.00 | 88.43±0.22 | 82.29±0.00 | 60.25±- | 83.86±0.05/88.82±0.02 |
| | | 50277 | -±- | -±- | -±- | -±- | -±-/-±- |
| | | 8192 | 77.33±0.00 | 85.73±0.14 | 84.04±0.00 | 60.51±- | 85.30±0.08/89.65±0.06 |
| None (Naive) | Word | 16384 | 77.53±0.00 | 87.92±0.19 | 84.29±0.00 | 62.18±- | 86.16±0.07/90.11±0.09 |
| | | 50277 | 78.34±0.00 | 91.04±0.29 | 84.29±0.00 | 65.15±- | 87.64±0.08/91.31±0.05 |

Table 14: Comparison of downstream performances across tokenizers in Hindi.

| Pre-tokenizer | Tokenizer | Vocab Size | Downstream Tasks | | | | |
|---|---|---|---|---|---|---|---|
| | | | **WaW**<br>Accuracy | **WaM**<br>Accuracy | **WaW**<br>Accuracy | **POS**<br>F1 | **POS-Pooled**<br>F1 |
| None (Naive) | Character | - | 54.18±0.57 | 73.54±1.36 | 58.78±1.46 | 66.36±0.54 | 41.96±0.36 |
| | | 8192 | 54.47±0.24 | 74.51±0.17 | 63.01±0.98 | 76.57±0.22 | 21.12±1.67 |
| None (Naive) | BPE | 16384 | 55.20±0.44 | 75.84±0.17 | 64.21±0.55 | 74.64±0.30 | 32.46±1.55 |
| | | 50277 | 55.49±0.14 | 74.32±1.29 | 63.46±0.34 | 73.86±0.94 | 42.13±0.19 |
| | | 8192 | 56.04±0.54 | 76.10±0.74 | 64.55±0.86 | 76.37±0.56 | 41.86±0.79 |
| Morfessor | BPE | 16384 | 55.31±0.25 | 75.84±0.81 | 62.78±0.78 | 74.30±0.50 | 41.36±1.62 |
| | | 50277 | 55.11±0.69 | 74.88±0.50 | 64.84±0.52 | 74.86±0.91 | 42.64±0.20 |
| | | 8192 | -±- | -±- | -±- | -±- | -±- |
| None (Naive) | Unigram | 16384 | -±- | -±- | -±- | -±- | -±- |
| | | 50277 | 66.64±0.38 | 80.82±0.29 | 68.27±1.57 | 94.34±0.11 | 34.52±0.16 |
| | | 8192 | -±- | -±- | -±- | -±- | -±- |
| Morfessor | Unigram | 16384 | -±- | -±- | -±- | -±- | -±- |
| | | 50277 | 60.13±0.53 | 82.05±0.22 | 67.75±1.24 | 93.37±0.01 | 34.70±0.09 |
| | | 8192 | 54.40±0.70 | 51.91±0.13 | 58.60±0.10 | 89.96±0.01 | 34.37±0.35 |
| Morfessor | Word | 16384 | 54.09±0.14 | 53.62±0.00 | 59.18±0.30 | 90.71±0.05 | 34.88±0.09 |
| | | 50277 | -±- | -±- | -±- | -±- | -±- |
| | | 8192 | 54.38±0.31 | 51.47±0.28 | 58.60±0.20 | 92.10±0.04 | 34.09±0.20 |
| None (Naive) | Word | 16384 | 54.24±0.52 | 55.15±1.17 | 60.26±0.26 | 93.21±0.04 | 33.92±0.03 |
| | | 50277 | 60.82±0.28 | 60.20±0.51 | 61.18±0.86 | 94.63±0.12 | 33.86±0.52 |

| Pre-tokenizer | Tokenizer | Vocab Size | Downstream Tasks | | | | |
|---|---|---|---|---|---|---|---|
| | | | **DM**<br>Accuracy | **STS-B**<br>Pearson/Spearman | **NER-CoNLL**<br>F1 | **SST-2**<br>Accuracy | **Dependency Parsing**<br>LAS/UAS |
| None (Naive) | Character | - | 55.99±0.39 | 16.05±3.40/14.76±3.62 | 57.44±0.25 | 71.10±1.40 | 21.68±1.82 |
| | | 8192 | 59.73±0.81 | 20.09±0.18/18.79±0.49 | 63.37±0.26 | 72.86±0.29 | 30.20±0.78 |
| None (Naive) | BPE | 16384 | 59.15±0.18 | 14.57±0.57/12.46±0.84 | 68.78±0.36 | 72.02±0.41 | 27.71±0.12 |
| | | 50277 | 57.22±0.48 | 14.14±0.72/11.96±0.46 | 67.38±0.31 | 71.86±0.63 | 29.73±0.08 |
| | | 8192 | 59.97±0.51 | 20.56±0.47/20.04±0.61 | 63.06±0.27 | 72.10±0.18 | 28.68±0.12 |
| Morfessor | BPE | 16384 | 58.28±0.18 | 21.17±2.83/20.21±3.12 | 68.05±0.11 | 71.75±0.76 | 27.97±0.08 |
| | | 50277 | 60.40±0.17 | 15.10±2.06/13.28±2.20 | 70.21±0.52 | 74.01±0.63 | 31.05±0.16 |
| | | 8192 | -±- | -±- | -±- | -±- | -±-/-±- |
| None (Naive) | Unigram | 16384 | -±- | -±- | -±- | -±- | -±-/-±- |
| | | 50277 | 62.10±0.29 | 49.83±0.74/48.39±1.03 | 91.37±0.08 | -±- | 69.60±0.32 |
| | | 8192 | -±- | -±- | -±- | -±- | -±-/-±- |
| Morfessor | Unigram | 16384 | -±- | -±- | -±- | -±- | -±-/-±- |
| | | 50277 | 61.93±0.31 | 22.97±0.63/22.09±1.08 | 89.06±0.15 | -±- | 64.88±0.07 |
| | | 8192 | 68.55±0.21 | 24.95±0.89/25.19±0.95 | 75.60±0.04 | 79.85±0.24 | 66.64±0.00 |
| Morfessor | Word | 16384 | 68.11±0.28 | 28.15±2.24/28.09±2.68 | 80.10±0.15 | 81.00±0.54 | 66.49±0.40 |
| | | 50277 | -±- | -±- | -±- | -±- | -±-/-±- |
| | | 8192 | 71.34±0.37 | 22.06±1.28/23.60±1.38 | 79.45±0.45 | 80.70±0.56 | 68.88±0.25 |
| None (Naive) | Word | 16384 | 72.01±0.32 | 27.53±1.77/27.48±1.91 | 85.35±0.22 | 83.94±0.23 | 70.96±0.05 |
| | | 50277 | 72.59±0.42 | 45.33±0.63/44.87±1.06 | 90.40±0.26 | 87.27±0.11 | 69.73±0.42 |

Table 15: Comparison of downstream performances across tokenizers in English.